\documentclass{ws-ijitdm}
\usepackage{tikz}
\usepackage{pgf-pie}
\usepackage{xcolor} 
\usepackage{bchart}
\usepackage{chronology}
\usetikzlibrary{trees}
\usepackage{graphicx,subfigure}
\usepackage[utf8]{inputenc}
\usepackage{subfigure}
\usepackage{pdfpages}
\usepackage{caption}
\begin{document}

\markboth{Imen JDEY, Ghazala HCINI and  Hela LTIFI}
{Deep learning and Machine learning for Malaria detection : Overview, challenges and future directions}

\catchline{0}{0}{0000}{}{}

\title{Deep learning and Machine learning for Malaria detection : Overview, challenges and future directions\\
}

\author{ Imen JDEY}
\address{Faculty of Sciences and Technology of Sidi Bouzid, \\
University of Kairouan, Kairouan, Tunisia 
}

\author{Ghazala HCINI}

\address{Faculty of Sciences and Technology of Sidi Bouzid, \\
University of Kairouan, Kairouan, Tunisia 
}
\author{Hela LTIFI}

\address{Faculty of Sciences and Technology of Sidi Bouzid, \\
University of Kairouan, Kairouan, Tunisia 
}
\maketitle

\begin{history}
\received{Day Month Year}
\revised{Day Month Year}
\end{history}

\begin{abstract}
 To have the greatest impact, public health initiatives must be made using evidence-based decision-making. Machine learning Algorithms are created to gather, store, process, and analyse data to provide knowledge and guide decisions. A crucial part of any surveillance system is image analysis. The communities of computer vision and machine learning has ended up curious about it as of late. This study uses a variety of machine learning and image processing approaches to detect and forecast the malarial illness. In our research, we discovered the potential of deep learning techniques as smart tools with broader applicability for malaria detection, which benefits physicians by assisting in the diagnosis of the condition. We examine the common confinements of deep learning for computer frameworks and organizing, counting need of preparing data, preparing overhead, real-time execution, and explain ability, and uncover future inquire about bearings focusing on these restrictions. 

\end{abstract}

\section{Introduction}	

\subsection{Study Background}
Malaria is one of the main global public health challenges \cite{1}, \cite{2}, \cite{3}. The origin of malaria begins from the landmass of Africa. The intestinal sickness was begun from the infection plasmodium falciparum infection which is the cause for this illness. The malady has travelled through all around the world by the pathogen of mosquitoes. The infection can survive in hot and gentle climate, but it cannot survive in exceptionally cold climate \cite{4}. In agreement with the World Health Organization (WHO), there were 228 million occurrences of the disease all over the world in 2019 \cite{5} \cite{6}. Malaria is caused by protozoan parasites of the genus Plasmodium transferred by bite of infected female Anopheles mosquitoes that infect red blood cells \cite{7}, \cite{8}, \cite{9}. The method of diagnosing malaria involves in a centrifuge classifying white blood cells and red blood cells so that only red blood cells can be used for analysis using a blood smear \cite{5}. It is a parasitic disease that causes severe contamination of red blood cells \cite{10}. It can be passed quickly through blood transfusions, syringes, and pregnant women to their babies \cite{9}. Malaria is a curable disease, but the lack of prompt and correct diagnosis and treatment can lead to serious health complications \cite{11}. It requires an early and accurate detection to control and eradicate this deadly disease \cite{11}. 
  
The Covid-19 widespread has affected malaria sickness conveyance in numerous settings for a few reasons. The efforts to control the widespread draw from existing constrained budgetary assets, clearing out other administrations underfunded. Moreover, reacting to pandemics requires an extra well-being workforce that was as of now rare. Due to constrained individual defensive gear (PPE) and testing capacity in a few settings, the accessible well-being suppliers are moreover likely to be uncovered and/or tainted. Ensuing separation, isolate, or some of the time passing encourage compounds the issue of well-being supplier shortage. Also, Covid-19 relief measures such as lockdown in nations may affect not as it were the worldwide and neighbourhood supply chain—including commodities fabricating, obtainment, shipping, and distribution—but moreover may affect malaria. 
For occasion, patients may not look for malaria sickness care due to constrained transportation alternatives or fear of contracting Covid-19. Patients may to look for care as well late, when extremely sick, and in this way at expanded hazard of mortality. Later considers assessed that the number of passings from malaria may twofold if the world’s consideration proceeds to be entirely on Covid-19 \cite{12}.\\
\begin{figure}[th]
\centering     
\subfigure[Figure A]{\label{fig:a}\includegraphics[width=60mm]{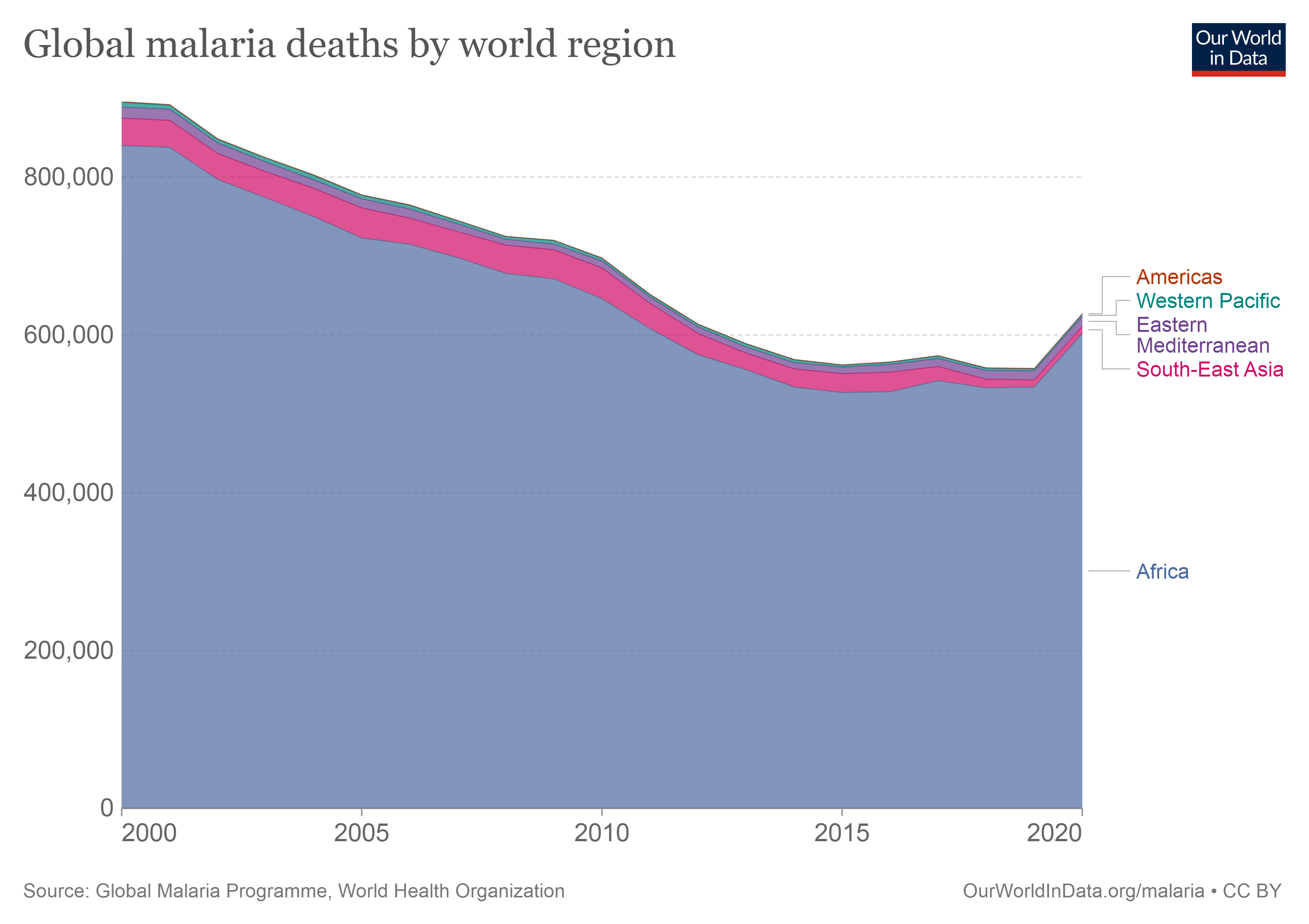}}
\subfigure[Figure B]{\label{fig:b}\includegraphics[width=60mm]{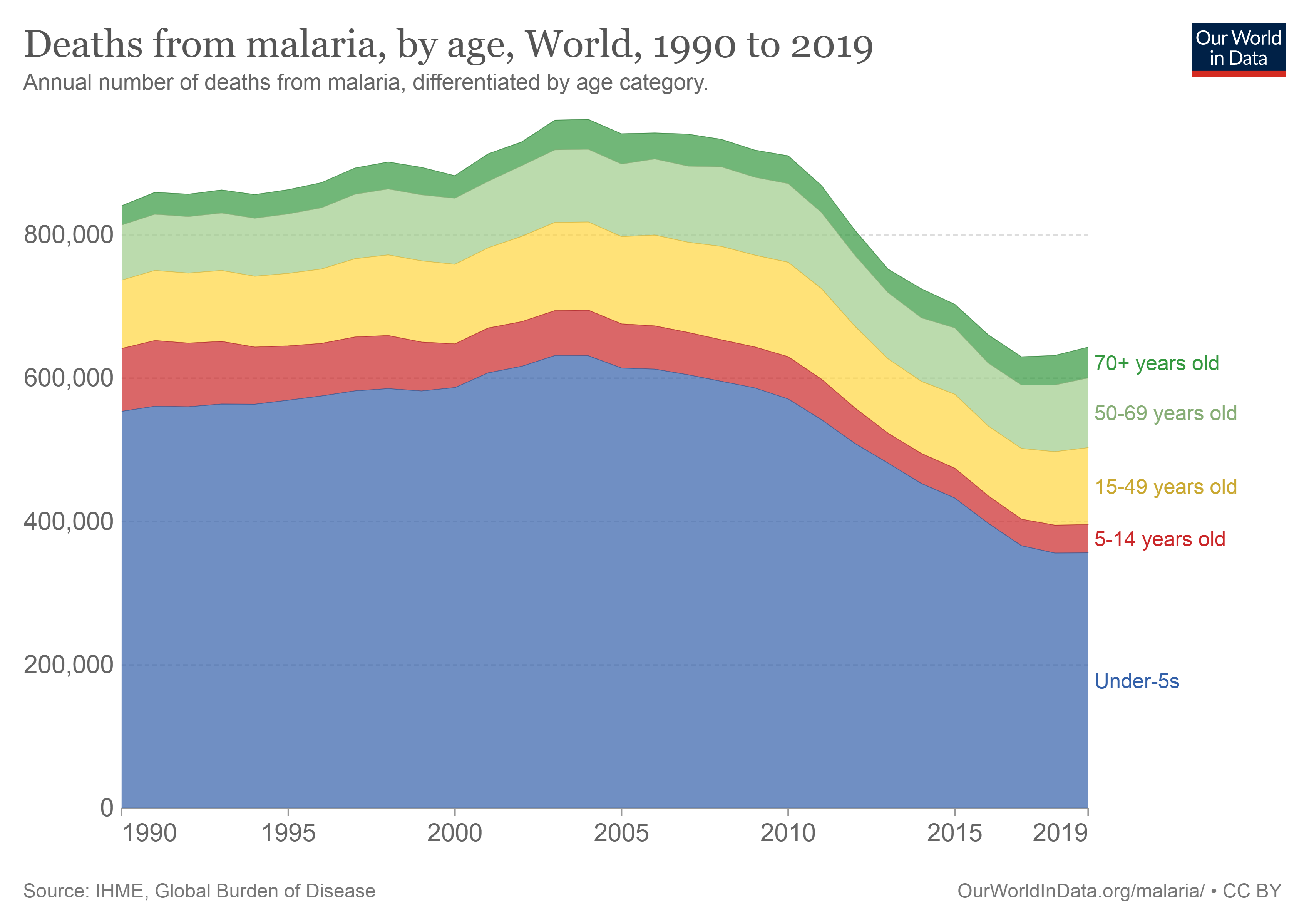}}
\caption{Malaria Death Rates}
\end{figure}
\newpage
If not appropriately treated, malaria can cause serious side effects such respiratory distress, pneumonia, anemia, and kidney failures. If a pregnant woman contracts the infection, stillbirths, neonatal deaths, and miscarriages result. The organs most impacted by malaria are the brain and lungs. Due to all of this, malaria is regarded as a medical emergency and needs to be detected right away. Delay in diagnosis causes fatalities. Finding malaria in infected patients is therefore crucial \cite{13}. In this paper, machine and deep learning will be used to identify malaria in its preliminary stages.
\subsection{Motivation}
We have noted that the scope of the consider ought to cover the nature of features representation and extraction, the achievability of different deep learning strategies, classification of malaria disease, benchmark datasets, reasonableness of the procedures in application settings, activity acknowledgment and peculiarity investigation yields, and assessment criteria in images. Our think about is spurred by a few viewpoints. Firstly, we stay on recognizing between conventional strategies based on manual highlights and those based on deep learning to highlight later propels in deep learning techniques for action recognition and malaria diagnosis in images. 
 Secondly, we distinguish issues amid image handlings. Too, we made a comparative study between the diverse strategies of machine learning and deep learning in arrange to have the foremost proficient. 
  \subsection{Overview of key concepts}
Machine Learning, Deep Learning and Convolutional Neural Network are all niche terms that are increasingly existing in scientific presentations. Machine learning is a field of Artificial Intelligence that attempts to create techniques capable of learning how to solve a specific task such as regression, classification, and clustering \cite{14}. 
Where the success of an approach is based on how accurately the algorithm can make predictions about datasets. 
\subsubsection{Machine Learning}
Machine learning (ML) has recently gotten a lot of press for its ability to properly forecast a wide range of complicated events. However, there is a growing recognition that, in addition to predictions, machine learning models may provide knowledge about domain relationships in data, which is referred to as interpretations \cite{15}.\\ 
ML trains computers to carry out tasks on their own by giving them instructions. It is a technique for data analysis that involves creating and fitting models and enables computers to "learn" via repetition and forecast future events. By analysing the content of the input image and extracting distinctive features from it, machine learning is used to diagnose malaria. These characteristics determine whether the prediction of the input image is given as a normal case or as an infected case. \\
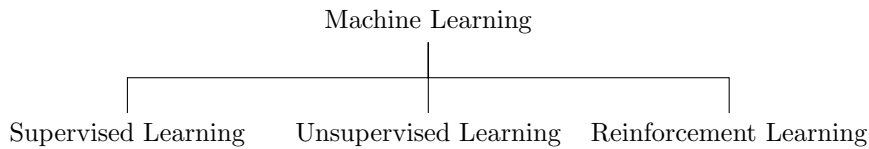
\begin{figure}[th]
\begin{center}
    \begin{tikzpicture}[level 1/.style={sibling distance=4cm},level 2/.style={sibling distance=2.5cm}]
	\node {Machine Learning}[edge from parent fork down]
		child { node {Supervised Learning}}
		child { node {Unsupervised Learning}}
		child { node {Reinforcement Learning}}
		;
\end{tikzpicture}
\caption{Types of Machine Learning}
    \label{fig:fig2}
\end{center}
\end{figure}

In the new millennium, reinforcement learning made a great amount of progress and now performs in several areas at a level comparable to that of a homo sapiens. Although model-based reinforcement learning is at the forefront of social robotics development, the medical field pays little attention to this idea. 
Most of the medical decision-making is sequential. Understanding the type of illness, a patient has required various test findings and a practical diagnosis session. Most machine learning models used in medicine frequently neglect the progressive progression of diseases, and doctors also have limited understanding of the nature of the conditional progression of a disease, except from their own experience \cite{16}. 

Unsupervised machine learning and supervised machine learning are the two main types of machine learning. Unsupervised machine learning draws inferences from datasets that contain input data without labelled replies, or in other words, where the expected output is not provided \cite{17}. The goal of supervised machine learning techniques is to discover the relationship between input variables (independent variables) and a target variable (dependent variable). Supervised approaches are further divided into two groups: classification and regression. The output variable in regression accepts continuous values, whereas the output variable in classification takes class labels \cite{18}. 
\paragraph{Regression}Regression is a method that is used to test two theories. For starters, regression analyses are commonly employed for forecasting and prediction, and their use overlaps heavily with machine learning. Second, regression analysis can be used to discover causal relationships between independent and dependent variables in particular situations. Importantly, regressions alone reveal only relationships between a dependent variable and a set of fixed variables in a dataset \cite{19}.
\paragraph{Clustering}
Clustering algorithms are unsupervised machine learning approaches that are frequently used when labels are unavailable (such as resource allocation problems) \cite{20}\\
\cite{21}. Clustering aids in the analysis of unstructured and high-dimensional data such as sequences, expressions, phrases, and images \cite{22}.
\paragraph{Classification}Classification is a data mining (machine learning) technique for predicting data instance group membership. Although there are a variety of machine learning approaches available, classification is the most extensively employed \cite{23}. In machine 
learning, classification is a well-liked task, particularly in future planning and knowledge discovery. Researchers in the disciplines of machine learning and data mining consider classification to be one of the most studied challenges \cite{18}.\\ 
One of the most fundamental challenges in medical image analysis is classification. Disease diagnosis and prognosis, anomaly identification, and survivorship prediction are typical medical image classification tasks. A label from one of the predefined classes would be assigned based on the information that was collected from the image. Since many classification techniques are completely supervised, a dearth of excellent labelled data frequently causes these techniques to fall short in actual clinical contexts \cite{24}.\\ 
\begin{figure}[th]
\begin{center}
\includegraphics[width=9cm, height=4.64cm]{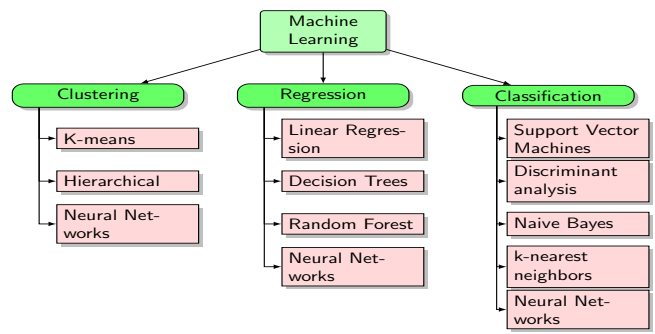}
\caption{The most popular algorithms for machine learning}
    \label{fig:fig3}
\end{center}
\end{figure}

Some studies chose to get these characteristics via a reliable technique like deep learning as machine learning techniques require extracted features to fulfil the classification tasks. Their models used CNN (Convolutional Neural Network) architectures to extract features, which were then input to one of the ML approaches. It is a combination of the CNN and ML blocks, with one being utilized for feature extraction and the other for classification \cite{25}.
\subsubsection{Deep Learning}
Faced with the exponential growth of data, a new field of machine learning has been used named Deep Learning \cite{17}. It is the latest trend in ML for its superior performance on big data \cite{14}. Recently, deep learning is increasingly addressed in computer aided diagnostic systems \cite{26}. They are widely used in medical image classification \cite{27} \cite{28}. Particularly, Convolutional Neural Networks (CNN), a class of Deep Learning models \cite{29}, have shown promising results in the tasks of classification, recognition, and localization of images \cite{30}\cite{31}. We are interested in image classification, the most important task in image processing \cite{32}. Most of the proposed methods are employed to solve the detection, segmentation, and classification tasks, as it has been clarified in figure 2 below \cite{33}. Multilayer neural networks (NNs) try to learn representations from input images, which allow us to achieve specific results such as segmentation or classification without handcrafted features \cite{34}. 

Areas of research involving deep learning architectures for image treatment have grown recently. One of the most popular deep neural networks architectures is the Convolutional Neural Network (CNN) \cite{35}. It is a supervised deep learning architecture. 
CNN has multiple layers; namely the convolutional layer, the non-linearity layer, the aggregation layer, and the fully connected layer \cite{36}. Most common CNN architectures are (AlexNet \cite{37}, VGG16, VGG19 \cite{38}, ResNet50, DenseNet201 \cite{39}, and Inception-v4) \cite{40}. To train these models of computer vision system, it is necessary to reach a considerable number and a wide variety of images \cite{38}, below the most popular architectures of CNN.\\ 
\begin{figure}[th]
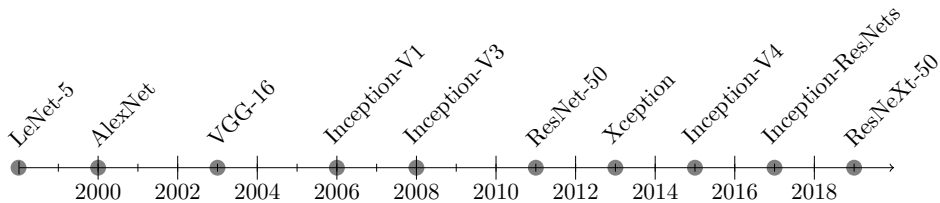

\begin{chronology}[2]{1998}{2019}{\textwidth}
         \event{1998}{LeNet-5}
         \event{2000}{AlexNet}
         \event{2003}{VGG-16}
         \event{2006}{Inception-V1}
         \event{2008}{Inception-V3}
         \event{2011}{ResNet-50}
         \event{2013}{Xception}
         \event{2015}{Inception-V4}
         \event{2017}{Inception-ResNets}
         \event{2019}{ResNeXt-50}
      \end{chronology}
\caption{CNN Architectures Timeline (1998-2019)} \label{fig1}
\end{figure}
\newpage
\begin{itemize}
    \item \textbf{AlexNet}: The first deep CNN (Convolutional Neural Network) architecture, AlexNet (Krizhevsky et al. 2012), is regarded as having produced ground-breaking outcomes for image classification and identification applications. Krizhevesky et al. proposed AlexNet, which deepened the CNN's learning capacity and applied several parameter optimization techniques \cite{40} \cite{41}. 
    \begin{figure}[th]
\begin{center}
\includegraphics[height=6cm, width=9cm]{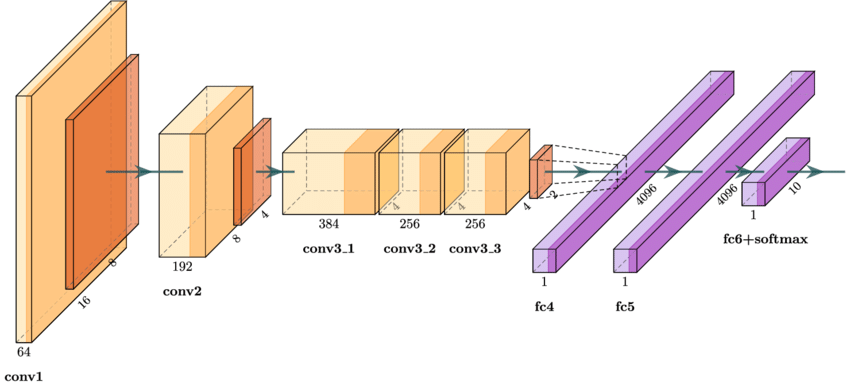}
\caption{AlexNet architecture }
    \label{fig:fig4}
\end{center}
\end{figure}
    \item \textbf{VGGNet (Visual Geometry Group)}: VGGNet was a neural network that did exceptionally well. It came in first place for image localization and second place for image classification \cite{42}\cite{44}, \cite{43}.
     \begin{figure}[th]
\begin{center}
\includegraphics[height=5cm, width=10cm]{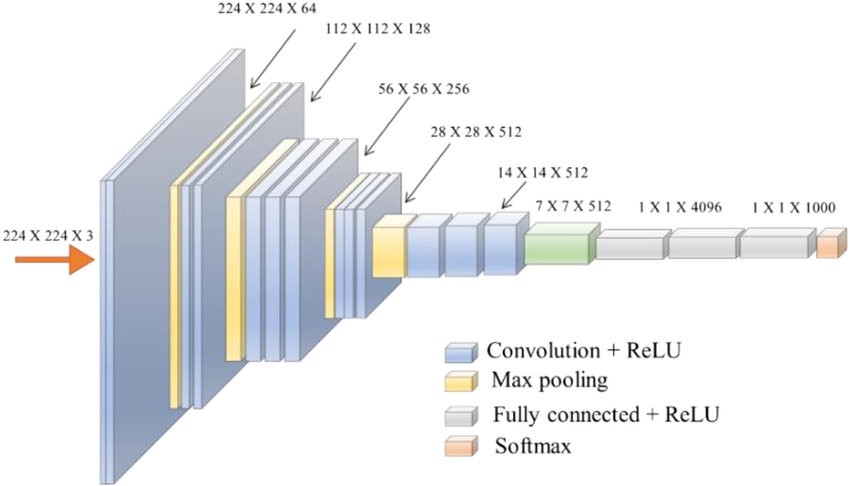}
\caption{VGGNet  architecture}
    \label{fig:fig5}
\end{center}
\end{figure}
\newpage
\item \textbf{ResNet (Residual Network)}: He et al. proposed ResNet, which is seen as a continuation of deep networks (He et al. 2015a). By introducing the idea of residual learning in CNNs and developing an effective approach for deep network training, ResNet altered the CNN architectural race.
\begin{figure}[th]
\begin{center}
\includegraphics[height=5cm, width=9cm]{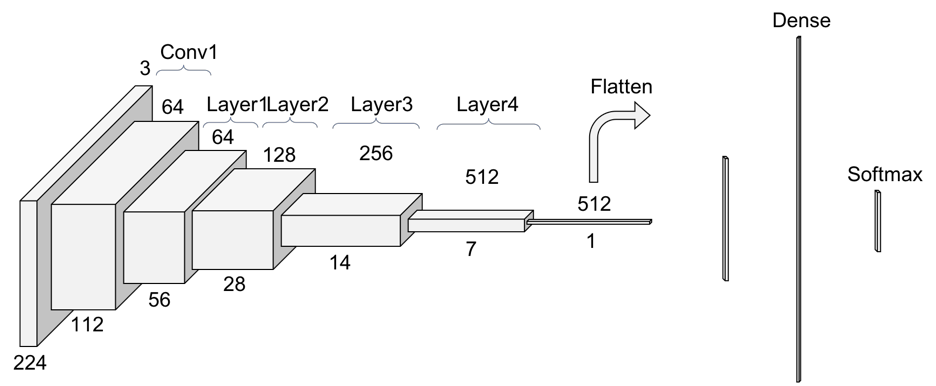}
\caption{ResNet architecture: ResNet 34}
    \label{fig:fig6}
\end{center}
\end{figure}
\item \textbf{DensNet (Densely Connected Network)}: Gao et al. created DenseNet in 2017 \cite{45}. The outputs of each layer relate to all successor levels in a dense block, which is made up of densely connected CNN layers. As a result, it has dense connectivity between the layers, earning it the moniker DenseNet. This idea is effective for feature reuse, resulting in a significant reduction in network parameters. DenseNet is made up of dense blocks and transition blocks that are positioned between two dense blocks \cite{46}. 
\begin{figure}[th]
\begin{center}
\includegraphics[height=4cm, width=9cm]{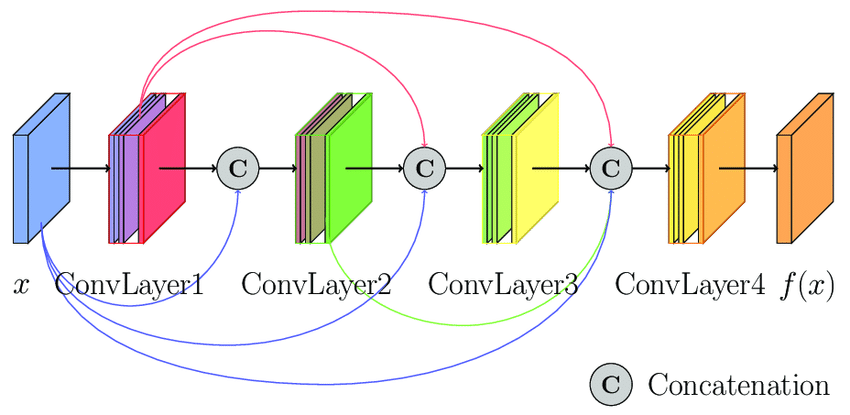}
\caption{An illustration of the DenseNet model}
    \label{fig:fig7}
\end{center}
\end{figure}
\newpage
\item \textbf{Inception}: The basic idea behind the Inception architecture is to investigate how a convolutional vision network’s best local sparse structure might be approximated and covered by conveniently available dense components. In general, an Inception network is made up of modules of the types stacked on top of each other, with occasional max-pooling layers of stride 2 to reduce the grid’s resolution \cite{49}\cite{50}.
\begin{figure}[th]
\begin{center}
\includegraphics[height=5cm, width=9cm]{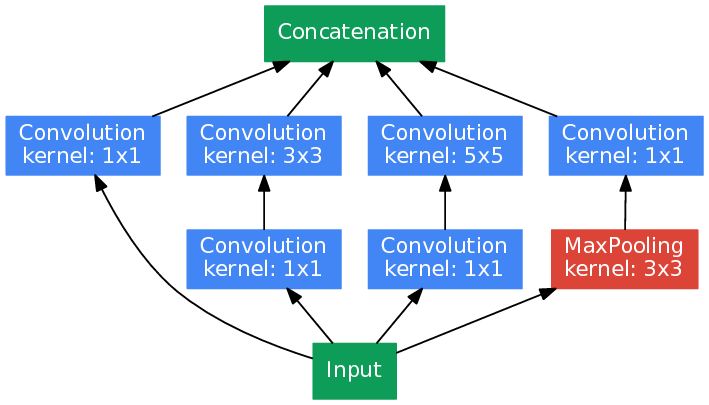}
\caption{Inception model }
    \label{fig:fig8}
\end{center}
\end{figure}
\end{itemize}

Unlike humans, deep-learning algorithms require a large library of high-quality, annotated data to learn and make accurate decisions about future events. This may be one of the reasons why the medical field has been slow to adopt the modern technology during its preliminary stages, as annotated training sets are difficult to come by and privacy concerns abound. Interestingly, using a technique called as transfer learning, the trained deep-learning models may be utilized to address issues in distinct purpose related applications. Deep learning has been used in the medical profession to solve problems including facial recognition \cite{51} \cite{52}, accurate skin burn classification \cite{53}, and cancer diagnosis \cite{54}, as well as in financial fraud detection. Surprisingly, a similar strategy was recently used to distinguish between 
blood-smear images with and without the Plasmodium parasite \cite{55}. 
\section{Malaria classification using machine learning}
The following search terms were used to find papers on artificial intelligence and malaria: "deep learning," "machine learning," "hybrid algorithms," "malaria," and "cell images".
\newpage
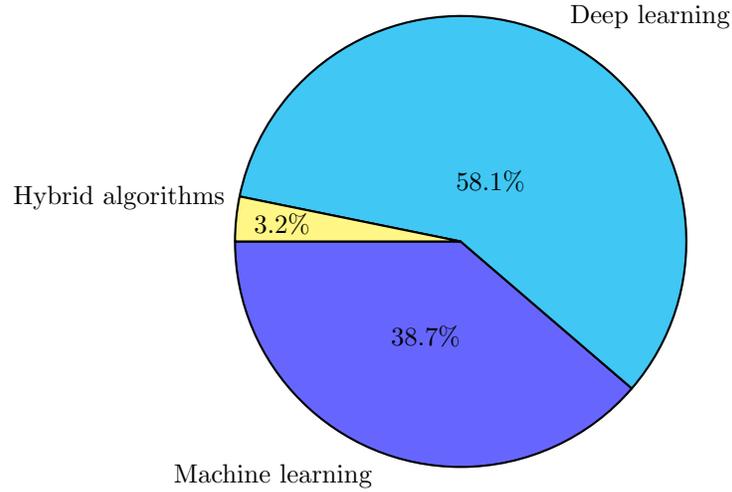
\begin{figure}[htbp]
\begin{center}
     \label{fig:my_label}
\begin{tikzpicture}
\pie [rotate = 180]
    {38.7/Machine learning,
   58.1/Deep learning, 3.2/Hybrid algorithms}
   \end{tikzpicture}
   \caption{Malaria diagnostic update: 2020-2022}
   \end{center}
\end{figure}

Recently, there have been a growing number of studies devoted to the application of machine learning technologies to automate and to accurate malaria diagnosis to save human life. We have selected several recent works from 2018 until 2022.
\subsection{Public Datasets}
Starting with explanations corresponding to different works used in the most famous and available databases (a public database containing 27,558 images) from \textit{https://ceb.nlm.nih.gov/repositories/malaria-datasets/}.
\begin{figure}[th]
\begin{center}
\includegraphics[height=5cm, width=10cm]{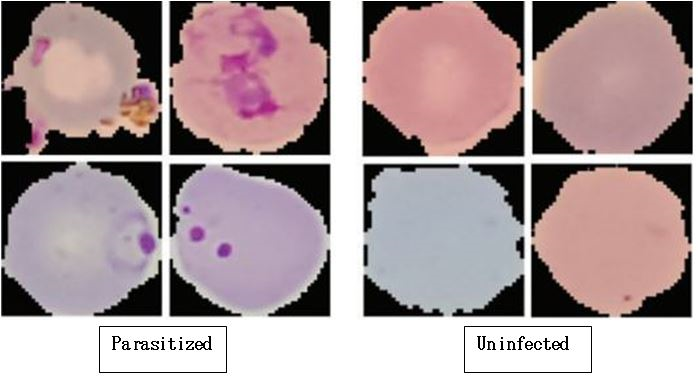}
\caption{Visualization of the dataset: parasitized and uninfected}
    \label{fig:fig9}
\end{center}
\end{figure}

Rajaraman et al. \cite{56} have proposed feature extractors using pre-trained CNN-based deep learning models for the classification of uninfected and parasitized blood cells to allow the identification of diseases. The proposed model has three convolutional layers which use 3 × 3 filters with 2-pixel steps, with 32 filters for the first and second convolutional layers and 64 filters for the third convolutional layer, and two fully connected layers. The resolution of the model input constitutes segmented cells is 100 × 100 × 3 pixels.  They evaluated the performance of pre-trained CNNs, specifically AlexNet, VGG-16, Xception, ResNet-50 and DenseNet-121 to extract characteristics of parasitized and uninfected cells. These models were optimized for hyper parameters with the Random Grid Search method. 

Masud, Mehedi, et al. \cite{57} proposed a CNN (Convolutional Neural Network) model composed of four convolutional blocks and two fully connected dense layers, applied to classify 27,558 segmented cell images, the given segmented cell images have a resolution of 224 × 224 × 3 pixels. The trainable parameters of their proposed model have a smaller size (409 K) compared to other complex models; its simplicity will be the first step while dealing with the problem of overfitting. 

In \cite{58}, Convolutional neural network models (InceptionV3 networks, GoogleNet, AlexNet, Resnet50, Vgg16) that have different depth architectures, were trained through BEKTAŞ, Jale by exploiting a cross validation by six were applied. the performance evaluation of these models was carried out via Support Vector Machine (SVM.). The choice of SVM is based on its performance because it is one of the most popular image recognition and classification machine learning algorithms \cite{59}. 

Alqudah, A. M., \& Qazan, S. \cite{60} proposed a new CNN model using transfer learning to classify red blood cells into two classes infected and Uninfected.

Umer, Muhammad, et al. \cite{61} implemented a novel model to detect malaria applied to 27, 558 cell images, collected from the National Institute of Health. Going through the main step which is pre-processing to set the image size of 120 × 120 pixels. In addition, they applied the normalization of the spots to preserve the characteristic image. The stacked model was designed from 22 layers in total, 5 convolution layers, 2 max-pooling layers, 4 dense layers, 1 mid-layer pooling, 1 flattened layer, 8 layers with 20\% abandonment and 1 fully connected layer. And as activation function, they used linear rectified (ReLU). To get a dimension of 120 × 120 from the first step, the image must be resembled by 200 × 200 pixels to make final decision, with Batch size value 32, continued for 13 epochs, and as optimizer they used ADAM. 

QANBAR, M. M., \& Tasdemir, S. \cite{62} developed a   deep   learning   model named Residence Attention Network (RAN). The residual attention network (RAN) is structured by combining several attention modules. Such as each attention module is divided into two branches: the mask branch and the trunk branch; the mask branch and the trunk branch. This approach applied on a dataset contained a total 27.558 cell images (20.658 images for training, 6.900 images for the test). 

Sinha, Shruti, et al. \cite{63} in his paper proposed a deep learning approach. This model uses two different components of CNN (Sequential and ResNet). Indeed, the difference between the two architectures is that ResNet uses more hidden layers rather than sequential ones. Applied on a dataset which contains 27,558 cell images with 80\% for training and 20\% for testing. 

Sandhya, Y., Sahoo, P. K., \& Eswaran, K. \cite{64} proposed an automated system based on deep learning the Convolutional Neural Network following these steps (1) Image Processing used to get the quality image then the application of CNN algorithm which contains the following layers; Convolution layer, Pooling layer, ReLU layer and fully connected layer. This model applied to classify 27,558 images with 30*30 pixels. 

Bhansali, R., \& Kumar, R. \cite{65} developed a deep learning approach. Starting with the first step Image Pre-processing then the use of CNN architecture which contains seven-layer (two convolution layers, two pooling layers, a flattening layer, and two fully connected dense layers). They used as optimizer Adaptive Moment Estimation (Adam). And to avoid the problem of overfitting networks, they carried out the early stopping, such as the number of parameters equal to 2,177,185 parameters were trained and optimized. 

In 2021, Irmak, E. \cite{66} proposed a novel CNN architecture designed to detect malaria from 27,558 thin blood cell images resized to 44 × 44 × 3. This model contained four convolution layers, four ReLU layers, four normalization layers, four max pooling layers, one fully connected layer, one Softmax layer, and one classification layer.

Engelhardt, E., \& Jäger, S. \cite{67} proposed 3 different model architectures. The first model was used to evaluate training loss and accuracy as the number of trainable parameters 1 212 513. The second sequential model has 9 841 025 trainable parameters, and the third model has 15 924 289 trainable parameters. 

Maqsood, Asma, et al. \cite{68} offered a customized CNN model (5 convolutional layers, 5 max-pooling layers, and 2 fully connected layers) that beats all deep learning models currently available. Data pre-processing, feature extraction, and classification are the three aspects of the suggested method. 

Elangovan, P., \& Nath, M. K. \cite{69} proposed a CNN model with four convolutional layers (C1, C2, C3, C4), two max pooling layers (P1, P2), and one fully connected layer (FC), using thin smear blood pictures to detect malaria parasites. Pre-processing, feature extraction, and classification are all stages of the proposed technique.

Çinar, A., \& Yildirim, M. \cite{70} used a deep learning algorithm to identify the presence or absence of malaria parasites, applied to dataset contains 3730 parasitic data and 3000 healthy data.  Without a Gaussian filter, the precision value is 94.33\%, More successful results were obtained with Gaussian filter applied data is value of 97.83\%. 

In malaria blood sample images of balanced class dataset generated from the Kaggle dataset, a novel deep learning model dubbed a data augmentation convolutional neural network (DACNN) was developed by Oyewola, David Opeoluwa, et al. \cite{71}.

Magotra, V., \& Rohil, M. K. \cite{72} presented a bespoke CNN with six convolutional layers and filter sizes ranging from 16 to 128 followed by an activation function called "Relu" and six max-pooling layers. We employed a dropout function three times at different points in the training process to reduce overfitting and make the model more durable in real-world applications. The custom CNN has a total of 332,577 parameters, including 332,161 trainable parameters (which are optimized and updated during training) and 416 nontrainable parameters (which are not optimized and updated during training). 
\subsection{Private Datasets}
Salamah, Umi, et al. \cite{73} proposed a robust segmentation technique going through the following stages of pre-processing, blood component segmentation, classification of blood components and parasite candidate formation. The performance was examined on 30 microscopic images of thick blood smears collected from various endemic places in Indonesia. While varying the cut off from 55 to 75 using 15 examples from each training and test data. 

PARDEDE, JASMAN, et al. \cite{74} introduced RetinaNet object detection approach for solving this problem. Data labelling is done with LabelImg in each erythrocyte object for normal and infected erythrocytes, subsequently saved in xml format, and converted to .csv file. In the training phase, they used Keras image data generator function like random flip, rotate and dilate to make preprocessing. Then, to extract the erythrocytes object they applied CNN model (the ResNet 101 and ResNet50) backend at 100 epochs. In the system test phase, pre-processing is first performed with each test image by clearing the ImageNet average. The RetinaNet architecture was applied to 25 images for testing and 2 images for training from ex vivo samples from P. vivax infected patients in Brazil [66], 75 images for testing and 4 images for training from Dr. Yani Triyani. 

Salamah, Umi, et al. \cite{75} proposed a method which is formed of    four    steps; (1) pre-processing (image   filtering and image enhancement), (2) Otsu thresholding (to separate the object and background), (3) determine region minima (this step is taken from a point which satisfies the condition, in which the collection point will establish a watershed.)  and (4) segmentation in which the performance of the proposed method is measured as a function of its accuracy value. 

Mustafa, Wan Azani, et al. \cite{76} implemented an image segmentation approach via morphological approach applied to processed images with dimension of 808 × 608 pixels and 24-bit depth prepared by Medical Microbiology \& Parasitology Department. It is a combination of filtering (mean filter and Gaussian filter) and morphological operator. 

In 2018, Roy, Kishor, et al. \cite{77} proposed as an approach to detect malaria parasite accurately a model based on two segmentation procedures that are watershed segmentation and HSV segmentation. The procedure HSV (Hue, Saturation and Value) is divided into multiple steps which are discussed below; (1) converting the image to HSV form and measuring parasite indices, (2) enhancing the Image, and (3) binary image conversion. Regarding Watershed Segmentation is used to solve the problem of HSV with it all the parasites cannot be detected because of some color distortion of the input images. Before the application of this technique, it is necessary to go through the following steps; (1) grayscale conversion, (2) morphological transformation, (3) adjustment of the image, (4) converting into black \& white image, (5) Finding the distance and nearest nonzero values, and watershed segmentation. 

In 2020, Pattanaik, P. A., Mittal, M., \& Khan, M. Z. \cite{78} constructed a computer-aided diagnosis (CAD) scheme used to identify the existence of malaria parasites in thick blood smear images, trained with a dataset of 1182 Field-stained malaria-infected blood smear microscopic images are collected from an Android smartphone at a Brunel SP150 microscope by a group of data scientists from Makerere University AI Research Group. The resolution of each image is 750 × 750 pixel. Three evaluation metrics highlighted to assess performance including K-Fold cross-validation, Class Performance (to measure the efficiency, they used the level of root mean square error and the ROC curve) and Baselines for Comparisons (accuracy and detection time). 

Adenowo, A. A., Awobajo, A. A., \& Alimi, S.  \cite{79} deployed a software-based diagnostic approach to detect Malaria Parasite in Blood. This system associates a set of image processing techniques like image acquisition, image pre-processing, image segmentation, feature extraction and classification applied to 92 Plasmodium parasite images were acquired in three categories: downloaded, captured, and digitally acquired images. 

Abraham, Julisa Bana. \cite{80} proposed an architecture of convolutional neural network named fully convolutional network approach. The architecture of U-Net is composed of two parts which are the contractive part and the expansive part. Concerning the contractive part, it is identical to the convolutional network architecture applied for the classification task. By accumulating unpadded 3 × 3 convolutional layers assisted by ReLU activation function and 2 × 2 max pooling with 2 strides for sub-sampling. Going to the second part the expansive part that has an oversampling accompanied by a 2 × 2 convolutional layer or layer signifies ascending convolution which split the entity layer in half, the ascending convolution is connected by two 3 × 3 convolutional layers and the activation function  ReLU. A layer called the dropout layer installed between the contractive and expansive layer as a regularization with a rate of 0.5 which aims to avoid overfitting of the model. The last layer is a 1x1 convolutional layer to map 64 feature vectors of the components to the expected output shape. U-Net architecture has a total of 23 convolutional layers, using Huber loss as a loss function. This model applied on 30 images of Plasmodium in thin blood smears with a size of 224×224 pixels. 

Sujith, Ch, et al. \cite{81} employed a CNN model for image classification task, which is a supervised learning problem.

Rosnelly, R., Wahyuni, L., \& Kusanti, J. \cite{82} proposed an optimized model for the development of the Malaria Parasite ROI (Region of Interest) process. Going through the following steps (1) Image Enhancement Results, (2) Binarization, (3) Region of Interest (ROI) by automatic cropping, and (4) ROI Image of Malaria Parasites.

Kurtuldu, Hüseyin, et al.\cite{83} implemented an intelligent system a hyperspectral microscope capable of measuring the intensity of light reflected or sent by illuminated materials in a wide range of wavelengths from visible to infrared, applied on a dataset of images were passed through a Gaussian filter with a 7x7 scan window to reduce noise. For the step of Classification, a supervised machine learning method was used to determine the optimal level of separation between the regions.

Kudisthalert, W., Pasupa, K., \& Tongsima, S. \cite{84} implemented a computer-aided automated diagnosis system. Red blood cell counts, and parasite life-cycle stage classification are the two fundamental components of the proposed framework. It is divided into eight steps: the first four were image pre-processing stages, the next two were RBCs segmentation stages, and the final two were classification tasks. 

Nugroho, Hanung Adi, et al. \cite{85} proposed an automated system for the detection of Plasmodium Ovale and Malaria Species on Microscopic Thin Blood Smear Images. Adaptive thresholding, color segmentation between the green (G) and hue (H) channels in the HSV color space, and morphological operation techniques were used. 

Madhu, Golla.\cite{86} developed a classification strategy using Extremely Randomized Trees classifier applied on a private dataset contained 400 images, through several steps such as image denoising using modified K-SVD, segmentation using fuzzy type-2 functions, feature extraction by local and global features, feature selection by Extra Trees Classifier, classification using Extra Trees Classifier, and diagnosis. 

We can see in the figure below the summary of different computer studies that have been conducted for malaria while emphasizing their performance. Image processing, machine learning and deep learning techniques are among them.\\
\begin{figure}[htbp]
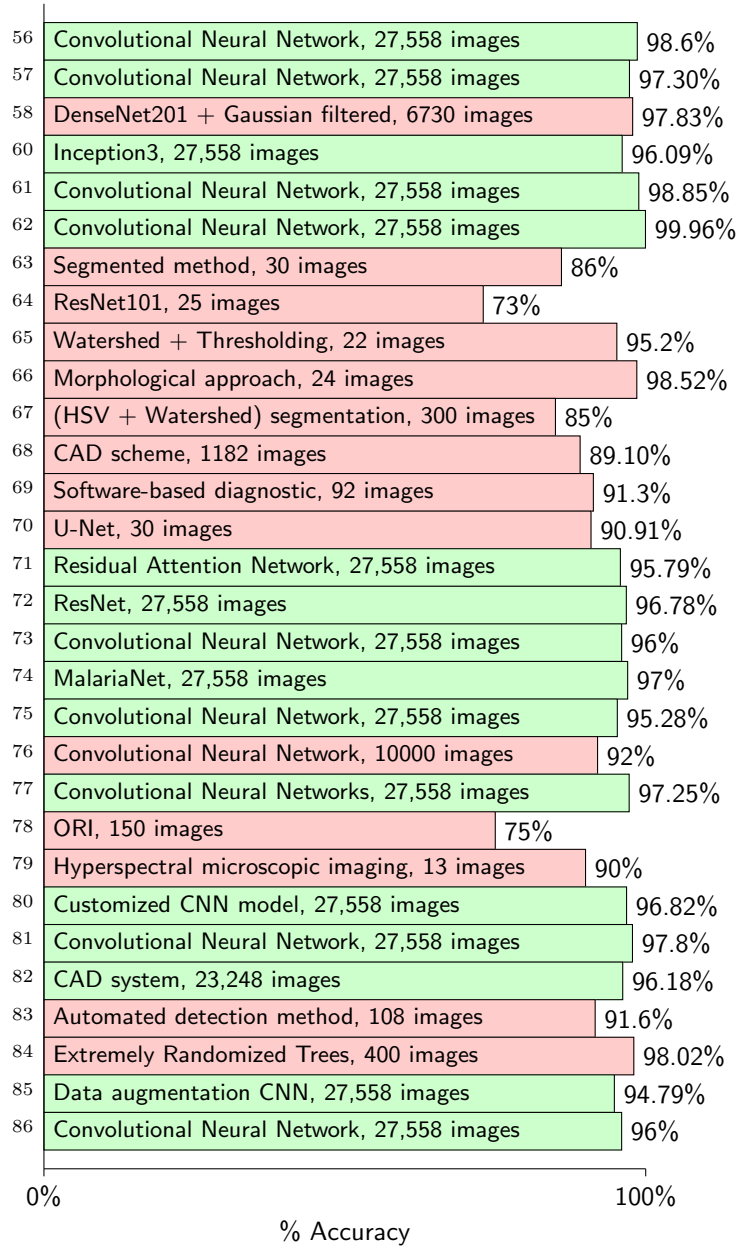

\begin{center}
    \label{fig:my_label}
\begin{bchart}[max=100, unit=\%]
 \bcbar[label={\cite{56}}, text=\small{Convolutional Neural Network, 27,558 images},
 color=green!20]{98.6}
 \bcbar[label={\cite{57}}, text=\small{Convolutional Neural Network, 27,558 images}, color=green!20]{97.30}
\bcbar[label={\cite{58}}, text=\small{DenseNet201 + Gaussian filtered, 6730 images}, color=red!20]{97.83}
 \bcbar[label={\cite{60}}, text=\small{Inception3, 27,558 images},
 color=green!20]{96.09}
  \bcbar[label={\cite{61}}, text=\small{Convolutional Neural Network, 27,558 images},
 color=green!20]{98.85}
  \bcbar[label={\cite{62}}, text=\small{Convolutional Neural Network, 27,558 images},
color=green!20]{99.96}
 \bcbar[label={\cite{63}}, text=\small{Segmented method, 30 images}, color=red!20]{86}
 \bcbar[label={\cite{64}}, text=\small{ResNet101, 25 images},
 color=red!20]{73}
 \bcbar[label={\cite{65}}, text=\small{Watershed + Thresholding, 22 images},
 color=red!20]{95.2}
 \bcbar[label={\cite{66}}, text=\small{Morphological approach, 24 images},
 color=red!20]{98.52}
 \bcbar[label={\cite{67}}, text=\small{(HSV + Watershed) segmentation, 300 images},
 color=red!20]{85}
 \bcbar[label={\cite{68}}, text=\small{CAD scheme, 1182 images},
 color=red!20]{89.10}
 \bcbar[label={\cite{69}}, text=\small{Software-based diagnostic, 92 images},
 color=red!20]{91.3}
 \bcbar[label={\cite{70}}, text=\small{U-Net, 30 images},
 color=red!20]{90.91}
 \bcbar[label={\cite{71}}, text=\small{Residual  Attention  Network, 27,558 images},
 color=green!20]{95.79}
 \bcbar[label={\cite{72}}, text=\small{ResNet, 27,558 images},
 color=green!20]{96.78}
\bcbar[label={\cite{73}}, text=\small{Convolutional Neural Network, 27,558 images},
 color=green!20]{96}
  \bcbar[label={\cite{74}}, text=\small{MalariaNet, 27,558 images},
 color=green!20]{97}
  \bcbar[label={\cite{75}}, text=\small{Convolutional Neural Network, 27,558 images},
 color=green!20]{95.28}
  \bcbar[label={\cite{76}}, text=\small{Convolutional Neural Network, 10000 images},
 color=red!20]{92}
  \bcbar[label={\cite{77}}, text=\small{Convolutional Neural Networks, 27,558 images},
 color=green!20]{97.25}
  \bcbar[label={\cite{78}}, text=\small{ORI, 150 images},
 color=red!20]{75}
\bcbar[label={\cite{79}}, text=\small{Hyperspectral microscopic imaging, 13 images},
 color=red!20]{90}
 \bcbar[label={\cite{80}}, text=\small{Customized CNN model, 27,558 images},
 color=green!20]{96.82}
 \bcbar[label={\cite{81}}, text=\small{Convolutional Neural Network, 27,558 images},
 color=green!20]{97.8}
 \bcbar[label={\cite{82}}, text=\small{CAD system, 23,248 images},
 color=green!20]{96.18}
 \bcbar[label={\cite{83}},  text=\small{Automated detection method, 108 images},
 color=red!20]{91.6}
 \bcbar[label={\cite{84}}, text=\small{Extremely Randomized Trees, 400 images},
 color=red!20]{98.02}
 \bcbar[label={\cite{85}}, text=\small{Data augmentation CNN, 27,558 images},
 color=green!20]{94.79}
  \bcbar[label={\cite{86}}, text=\small{Convolutional Neural Network, 27,558 images} ,
 color=green!20]{96}
 \bcxlabel{\% Accuracy}
 \end{bchart}
 \caption{Malaria diagnosis using deep learning 2018-2022, {\color{red!60} Private datasets}, \color{green!60}, Public datasets}
 \end{center}
\end{figure}
\newpage
\section{Hybrid Machine Learning Models for Malaria Classification}
The challenge of selection (and thus of understanding) and the developing issues of robustness lead to the got to reconsider the artificial intelligence approaches to the straightforwardness, interpretability or explainability of fake insights calculations. As well as the require for secure brilliantly frameworks.

In response to the increased demand for explainability, new algorithms are developing that offer precision, explainability, speed, and optimization of the amount of calculation required.\\ 
The hybrid model, which uses both machine learning techniques, produces outcomes that are better optimized \cite{87}. Recently, some research on fusing machine learning techniques into diverse hybrid approaches has become known. Further research is needed because this study is still in its initial stages \cite{88}. 

For the image classification of malaria cells, Diker, A. \cite{89} built a machine learning model based on Deep Residual CNN features Bayesian optimization method. The malaria cell images were classified using residual CNN, k-Nearest Neighbors (k-NN), and Support Vector Machine (SVM) classifiers. There are 27558 images of malaria cells in the public database that the National Institutes of Health (NIH) maintains. 

 A deep learning model using Canonical Correlation Analysis was proposed by Patil, A., et al \cite{90}. The CCA approach extracts, learns, and trains many nuclei patches at once to view the impacts of overlapping nuclei. Because blood cell pictures overlap, classification time is shortened, input image dimensions are compressed, and the network converges more quickly with more precise weight values. 12,442 blood cell images make up the dataset, with 9957 serving as training images and 2487 serving as test images. Eosinophil, Lymphocyte, Monocyte, and Neutrophil are the four different blood cell classes into which these picture samples are categorized. Compared to the testing dataset, which has 623 Eosinophil, 623 Lymphocyte, 620 Monocyte, and 624 Neutrophil photos, the training dataset contains 2496 Eosinophil, 2484 Lymphocyte, 2477 Monocyte, and 2498 Neutrophil images. Initially, RGB images with dimensions of 320 x 240 x 3 were used by the entire neural network.

  A malaria disease detection algorithm was put forth by Soomro, Aqeela, et al. \cite{91}. Along with the Naive Bayes Classifier, they applied image processing filters and methods to detect the malaria disease. The dataset used in their research is one of 27,558 images from the Kaggle online datasets pool. Images in the dataset range in size, thus it is important to resize them all to the same standard size of 300x300.

  To prevent inaccurate results brought on by human error, Marada Amrutha Reddy et al. \cite{92} developed a transfer learning model using inception-v3 and an SVM classifier applied to 27,560 cell images.

  A model was created by H. Mohamed, E. et al. \cite{93} that combines the deep models' advantage of automatically extracting features with the standard machine learning classifiers' greater classification accuracy. Logistic regression was used as a classifier and the pre-trained model MobileNet-224 as a feature extractor on 9,760 images.

A color-texture feature (YCrCb HOG) was introduced by M. K. Bashar \cite{94} using an unbalanced dataset of 46,978 single-cell thin blood smear images and an SVM classifier.

VGGNet was used by Sahlol, A. T. et al. \cite{95} to extract characteristics from WBC images. A statistically enhanced Salp Swarm Algorithm (SESSA) is then used to filter the retrieved features on 50,000 images for validation and 100,000 images for testing.

The Salp Swarm Algorithm and the Cat Swarm Optimization (SSPSO) algorithm-based optimized Convolutional neural networks (SSPSO-CNN) approach were combined, according to Kumar, R. et al. \cite{96}. Five types of peripheral blood cells, including eosinophils, basophils, lymphocytes, monocytes, and neutrophils, were categorized by it in the past. Their approach outlines the process of developing a classifier that has been fine-tuned using 10,674 images gathered from medical practice.

An algorithm based on the pretrained Alexnet and Google Net designs was proposed by A. Nar et al. \cite{97}. Both CNN designs' final pooling layer's feature vector has been combined, and the resulting feature vector is classified by the SVM (Google Net-SVM). 

Jha, K. K. et al. \cite{98} created a leukemia detection module using deep learning and blood smear images. Pre-processing, segmentation, feature extraction, and classification are performed during the detection phase. The segmentation outcomes of the active contour model and fuzzy C means method are combined in the suggested hybrid model based on mutual information (MI). The statistical and Local Directional Pattern (LDP) features are then retrieved from the segmented pictures and fed into the suggested Chronological Sine Cosine Algorithm (SCA) based Deep CNN classifier.

A two-step process was used by Abbas, S. S., et al. \cite{99}, beginning with image segmentation and ending with life stage classification. To segment and label 3000 images of Plasmodium falciparum parasites, scientists used a random forest classifier to determine whether each pixel was a parasite or not.

\begin{figure}[h]
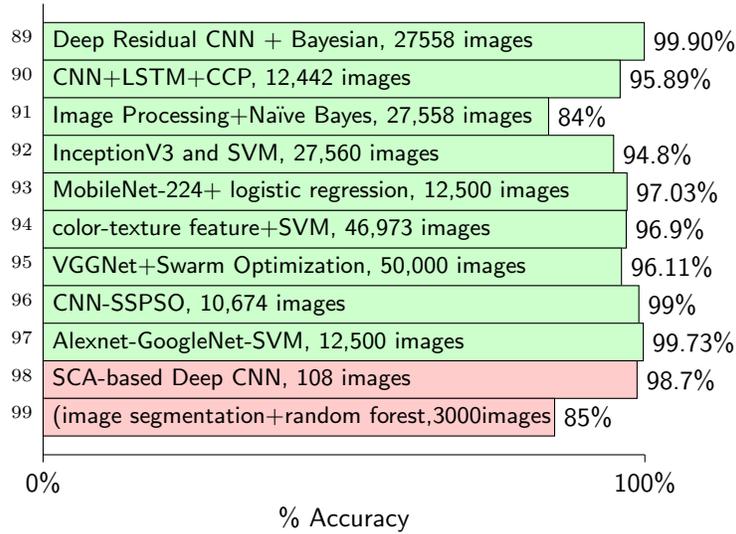

\begin{center}
        \label{fig:my_label}
\begin{bchart}[max=100, unit=\%]
 \bcbar[label={\cite{89}}, text=\small{Deep Residual CNN + Bayesian, 27558 images}, color=green!20]{99.90}
 \bcbar[label={\cite{90}}, text=\small{CNN+LSTM+CCP, 12,442 images}, color=green!20]{95.89}
\bcbar[label={\cite{91}}, text=\small{Image Processing+Naïve Bayes, 27,558 images}, color=green!20]{84}
 \bcbar[label={\cite{92}}, text=\small{InceptionV3 and SVM, 27,560 images},
 color=green!20]{94.8}
  \bcbar[label={\cite{93}}, text=\small{MobileNet-224+ logistic regression, 12,500 images},
 color=green!20]{97.03}
  \bcbar[label={\cite{94}}, text=\small{color-texture
feature+SVM, 46,973 images},
color=green!20]{96.9}
 \bcbar[label={\cite{95}}, text=\small{VGGNet+Swarm Optimization, 50,000 images}, color=green!20]{96.11}
 \bcbar[label={\cite{96}}, text=\small{CNN-SSPSO, 10,674 images},
 color=green!20]{99}
 \bcbar[label={\cite{97}}, text=\small{Alexnet‑GoogleNet‑SVM, 12,500 images},
 color=green!20]{99.73}
 \bcbar[label={\cite{98}}, text=\small{SCA-based Deep CNN, 108 images},
 color=red!20]{98.7}
 \bcbar[label={\cite{99}}, text=\small{(image segmentation+random forest,3000images},
 color=red!20]{85}
\bcxlabel{\% Accuracy}
 \end{bchart}
 \caption{Hybrid Machine Learning Models for Malaria Classification 2019-2022, {\color{red!60} Private datasets}, \color{green!60} Public datasets}
 \end{center}
\end{figure}
\newpage
\begin{table}[ph]
\tbl{Available datasets}
{\begin{tabular}{@{}ccc@{}} \toprule
Datasets & Number of images & Dataset link \\
 \\ \colrule
Malaria Cell images & 27,558 images & \textbf{Binary Classification}: \textit{https://ceb.nlm.nih.gov/proj/malaria/cell images.zip}\\
Malaria Cell images & 27,558 images & \textbf{Binary Classification}: \textit{https://www.kaggle.com/datasets/iarunava/cell-images-for-detecting-malaria}\\
Blood cells images& 12444 images & \textbf{Multi Classification}: \textit{https://www.kaggle.com/paultimothymooney/blood}\\
 \botrule
\end{tabular}}
\end{table}
\section{Discussion}
With the assistance of deep neural network systems, the demonstrative capabilities of learning calculations come closer to the level of human encounter, changing the CAD worldview from a "moment conclusion" instrument to a more collaborative utility. In this inquire about, we methodically analysed and summarized the distinctive strategies based on deep learning for the discovery and conclusion of malaria. Deep learning has accomplished gigantic triumphs in several distinctive areas; in any case, its genuine potential in medical imaging has however to be accomplished. Malaria detection and diagnosis using deep learning approaches have interesting challenges that must be tended to earlier to clinical appropriation. 
\subsection{Clinical and research engineers' communication gaps}
Miscommunication between the clinical and research expertise and data handling is a fundamental problem. Maintaining a positive working connection between clinical and research engineers throughout the research phase can boost the likelihood of success. Clinical professionals can provide suggestions on clinical settings using modern technologies, and a group of computer engineers can interpret study findings in innovative ways using their specialized knowledge. To create functional images of specific patients, researchers can combine the anatomical data acquired with innovative imaging technologies to computer simulations. Malaria engineering and diagnostics would result in cross-disciplinary collaboration with clinical management or marketing professionals, which could be beneficial. 
\subsection{Obtaining labeled data}
We must consider criteria in method applications such as the data delivered to us for the application, as they will clarify the decision making on the techniques to be used. Obtaining labelled data has significantly increased the potential to use deep learning models for their delivered performance. Labelling is done by several experts to avoid errors, noise, and missing values. So, to achieve a successful supervised learning method, it requires human (experts) annotated data. 
\subsection{Scarcity of imaging data in big data area}
An important consideration, Data Dependent, is that deep learning models work well on large data sets. Indeed, training and testing these models on a small dataset becomes impractical, so they are unable to learn from limited datasets. Therefore, these methods could prove to be effective with more training data. 

According to several studies, data augmentation will be a solution to having a large amount of data. However, it may introduce noise into the images or necessitate the sampling of overlapping image patches. However, the augmented samples can be highly associated with one another, leading to overfitting. Overfitting is a common machine learning and statistical modelling problem in which a learning model memorizes the training data and is unable to generalize to new data. Furthermore, local patches are unable to include the image's global and spatial context, resulting in mistakes. 
\subsection{Public dataset (Benchmark: Standard reference)}
If this study takes the route of a comparison of methods, to decide which is the most effective, it is suggested to compare these techniques by applying it to public databases and not to private ones collected from clinics and hospitals. A hassle associated with the datasets accrued from hospitals which cannot be used as input for deep learning techniques. Leveraging data from other hospitals can be challenging due to institutional differences with patients and data coding and collection. So, these models are not flexible. Here we are talking about normalization problem.  The main conclusion from this consideration is that Deep Learning models are not generalized. 
\subsection{Explainability of deep learning models}
In the two main phases of treatment; the learning phase and the testing phase, it is important to consider the complexity and time trade-offs that some methodologies radically include. A challenge associated with the use of deep learning models in medicine is that they use many parameters, which makes it difficult to determine the specific variables that give rise to predictions and worsens the traditional understanding of model overfitting. More complicated models are almost impossible to understand and explain.  Often Machine learning algorithms can be used to make decisions about a situation, but they are unable to explain what factors influenced the algorithm's decision. Most Machine learning and deep learning methods lack the ability to explain. For these reasons, control will be extremely limited over the Deep Learning algorithms. The main objective is to achieve simplification. Some studies suggested the combination of filtering (mean filter and Gaussian filter) to reduce noise. 
\subsection{What would it take for radiologists to accept deep learning tools in their everyday practice?}
Radiologists should be able to upload new data into the training system using real-world apps. However, it is unclear how each piece of uploaded data may be trained again, either from scratch or as a pre-trained network, and how training time and overfitting can be dealt with by regular training with slightly updated data sets. Finally, transferring these systems to a virtual cloud environment and making them available always would be a helpful feature. 
\section{Conclusion and Outlook}
Machine learning and deep learning are significant technologies that have changed the medical history. These paradigms have significantly changed how people store, process, and analyse data to make better decision. These have also led to research in developing technology that changes drastically with time. Machine learning techniques can be used to solve complicated real-world problems. The relevance of machine learning especially Deep learning in resolving the difficult issues of malaria detection and classification is highlighted in this article. This paper has presented a detailed review of all these new paradigms by particularly illustrating their contributions to the healthcare system. We provided some details about the different architectures of deep learning especially convolutional neural network. Also, we presented challenges that militate against the performance of Deep learning models. We will strongly recommend in this paper that artificial intelligence-based drug program primarily deep learning would significantly benefit in the fight against malaria in terms of saving human lives in less time and less cost. 

In future studies, for best results, a pre-trained algorithm customized and modified on a dataset benchmark may need to be developed to reduce the complexity of the default models.  
Clinicians' performance on the newly created system (CAD) should be compared to their performance on an existing tool, if appropriate, for clinical assessment. We expect to see more CAD systems for malaria detection and diagnosis as the adoption of deep learning algorithms, hardware support, and compelling qualities of deep learning methods such as robustness, generalizability, and expert-level accuracies rise.

\appendix

\end{document}